\documentclass[10pt,twocolumn,letterpaper]{article}
\usepackage[pagenumbers]{cvpr}
\usepackage{booktabs,multirow}

\usepackage{cuted}
\usepackage{tabularx}
\usepackage{caption}
\usepackage{booktabs,amssymb,colortbl}
\definecolor{ablationgray}{gray}{0.94}
\definecolor{cvprblue}{rgb}{0.21,0.49,0.74}
\usepackage[pagebackref,breaklinks,colorlinks,allcolors=cvprblue]{hyperref}

\def\confName{CVPR}
\def\confYear{2027}

\title{\LARGE CAST: Calibrating Correspondences for Training-Free\\Biomedical In-Context Segmentation}

\author{
Mingze Jiang$^{1}$ \quad
Jiashu Han$^{1}$ \quad
Haonan Chen$^{2}$ \quad
Zhiyi Li$^{1}$ \quad
Yilun Du$^{2}$ \quad
Kunzan Liu$^{1,*}$\\[0.4em]
{\normalsize $^{1}$Massachusetts Institute of Technology \qquad $^{2}$Harvard University}
}

\begin{document}
\maketitle
{\renewcommand{\thefootnote}{\fnsymbol{footnote}}%
\footnotetext[1]{\href{mailto:kunzan@mit.edu}{\texttt{kunzan@mit.edu}}}}

\begin{strip}
    \vspace*{-3.6em}
    \centering
    \includegraphics[width=\linewidth]{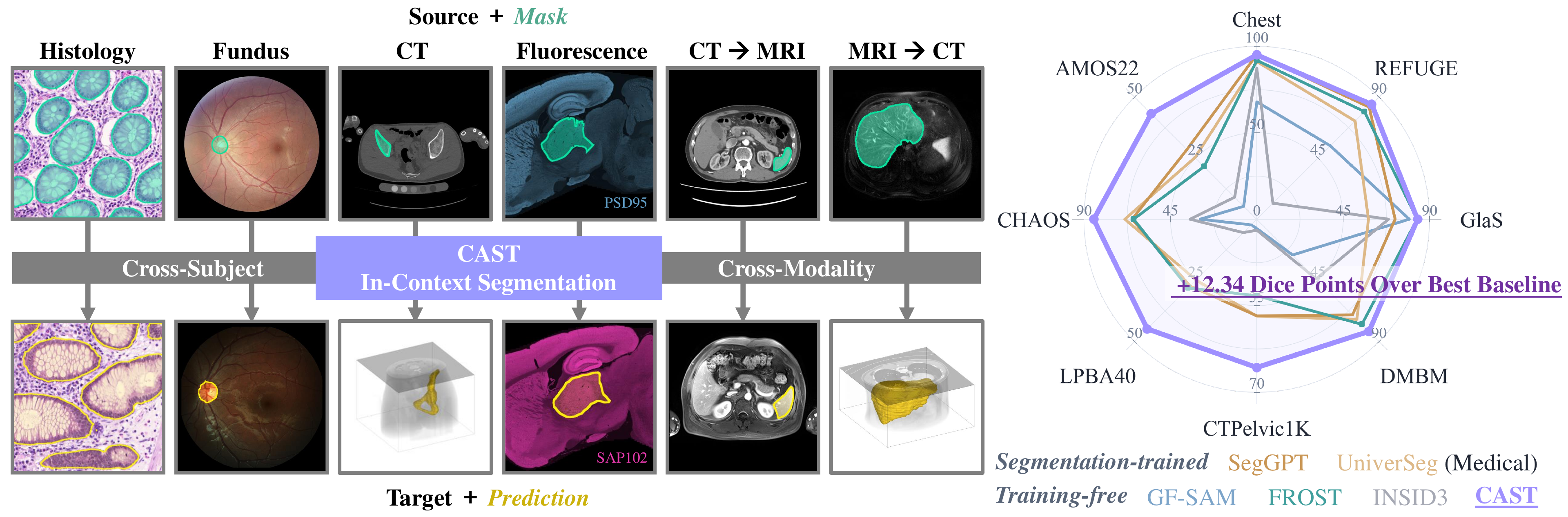}
    \captionsetup{hypcap=false}
    \captionof{figure}{
    \textbf{Overview of CAST for training-free biomedical in-context segmentation.}
    Left: Representative source--query examples spanning
    cross-subject, cross-marker, and cross-modality transfer.
    Green and yellow contours indicate source annotations
    and query predictions, respectively.
    Right: Dice comparison across eight biomedical benchmarks.
    CAST improves the overall average by 12.34 percentage
    points over the strongest baseline.
    }
    \captionsetup{hypcap=true}
    \label{fig:overview}
\end{strip}

\begin{abstract}

Biomedical image segmentation must keep up with new
anatomical structures, subjects, imaging modalities,
and acquisition protocols, yet dense expert annotations
are scarce and models trained on one cohort often
struggle to transfer.
In-context segmentation (ICS) offers a way out:
a new structure is specified by a few annotated exemplars
at inference time.
However, reliable correspondence remains challenging
in biomedical images, where adjacent structures often
share similar appearances, and variations across
subjects and imaging conditions further obscure
their distinctions.
We introduce \textbf{CAST},
a training-free framework built on a frozen DINOv3 backbone
to calibrate fine-grained correspondences.
Its stability test improves feature reliability at
both global and patch-specific levels, while its
coherence test assesses whether candidate
correspondences are supported by consistent local
structural context.
Together, these tests enable more reliable correspondence
across diverse anatomical structures and imaging conditions.
Across eight biomedical benchmarks spanning 2D and 3D
segmentation, cross-subject, and cross-modality
transfer, CAST outperforms the strongest overall baseline
by 12.34 percentage points in Dice and ranks first across
all six volumetric settings.
The versatility of CAST further opens up opportunities
for challenging biomedical applications, including
2D-to-3D segmentation and atlas label transfer.

\end{abstract}

\section{Introduction}
\label{sec:introduction}

Biomedical image segmentation has become remarkably
accurate and remarkably brittle.
Task-specific models encode anatomical priors and
domain expertise~\cite{isensee2021nnunet,wasserthal2023totalsegmentator},
but models trained on a limited source cohort can
struggle with shifts in scanners, acquisition protocols,
patient populations, and imaging modalities
~\cite{zhang2020stacked,guan2022domainadaptation,ouyang2023causality}.
Extending these models to new settings often requires
dense expert annotations and retraining, which are
costly and, for many specialized biomedical tasks,
difficult to obtain.
The field needs segmentation that adapts as fast
as the data changes.

In-context segmentation (ICS) offers that adaptability
by specifying the task rather than training for it:
a few annotated exemplars presented at inference time
define what to segment in a new image.
The paradigm has shown strong potential in natural
images~\cite{wang2023seggpt,wang2023painter}
and has inspired exemplar- and context-conditioned
approaches in medical imaging
~\cite{butoi2023universeg,rakic2024tyche,gao2025iris}.
Recent advances in vision foundation models have
further expanded its potential.
By leveraging the transferable representations of
frozen self-supervised vision transformers
~\cite{caron2021dino,oquab2024dinov2,simeoni2026dinov3},
training-free methods~\cite{cuttano2026insid3,hu2026foris,park2026frost}
have demonstrated promising results through direct
feature correspondence, without task-specific optimization.
This offers a flexible route to segmenting previously
unseen structures, but places a premium on reliable
matching between exemplar and query patches in
the frozen feature space.

\begin{figure}[t]
    \centering
    \includegraphics[width=0.95\columnwidth]{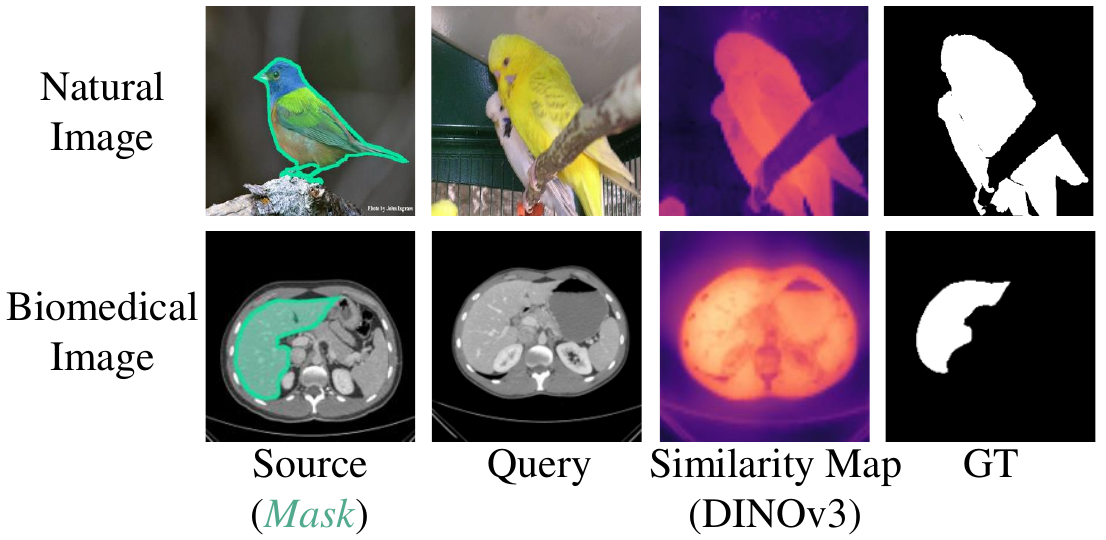}
    \caption{
    Comparison of DINOv3 feature similarity between natural
    and biomedical images.
    Similarity maps are obtained by computing cosine
    similarities between query patch features and annotated
    target features from the source image.
    Natural images exhibit clear similarity contrast
    between foreground and background, whereas
    biomedical images show similar responses across
    neighboring anatomical structures.
    Green contours indicate source annotations.
    }
    \label{fig:natural_biomed_similarity}
\end{figure}

We argue that this matching problem is particularly
challenging in biomedical images.
In many natural-image settings, target objects
exhibit clear appearance differences from
non-target regions.
Biomedical images, however, often contain multiple
adjacent anatomical structures that share tissue
appearance, intensity, and texture, with boundaries
defined by anatomical organization rather than
pronounced visual transitions, as illustrated in
\cref{fig:natural_biomed_similarity}.
The most challenging non-target regions in biomedical
images are often other anatomical structures.
We expect two consequences in frozen feature space.
First, subtle distinctions between neighboring structures
can be obscured by variations induced by geometry,
subject, and acquisition, allowing visually similar
but anatomically incorrect patches to rank above
the correct correspondences.
Second, non-target surroundings are not merely regions
to be rejected, but structured anatomy that provides
additional evidence for distinguishing visually similar
targets.
Consequently, center-level appearance similarity alone
is insufficient for reliable anatomical correspondence.

If appearance alone cannot be trusted, reliable
correspondences must draw on evidence beyond
individual patch similarity.
Our observation is that annotated exemplars provide
two complementary sources of such evidence:
the stability of visual features under geometric changes
and the spatial organization of labeled anatomy.
We introduce CAST, a training-free framework that
brings these sources together through two tests.
The stability test accounts for feature variations
under geometric transformations, while the coherence
test examines whether candidate matches are consistent
with their surrounding anatomical context.
By combining these complementary cues, CAST establishes
more reliable correspondences between exemplars
and query images, supporting segmentation across
diverse anatomical structures and imaging conditions.
Built upon the transferable representations of a
frozen DINOv3 backbone~\cite{simeoni2026dinov3},
CAST requires no additional training or
gradient-based adaptation.

We evaluate CAST on eight biomedical benchmarks
spanning 2D and 3D segmentation, cross-subject, and cross-modality transfer.
CAST achieves the highest overall performance,
exceeding the strongest baseline by 12.34 percentage
points in Dice and ranking first across all six volumetric
settings.
We further explore atlas label transfer, an important
step toward bringing standardized anatomical
knowledge to specimen-specific imaging.
CAST maintains promising performance in this
challenging setting, highlighting its potential
for adapting to diverse biomedical imaging scenarios.

\begin{figure*}[t]
    \centering
    \includegraphics[width=0.9\textwidth]{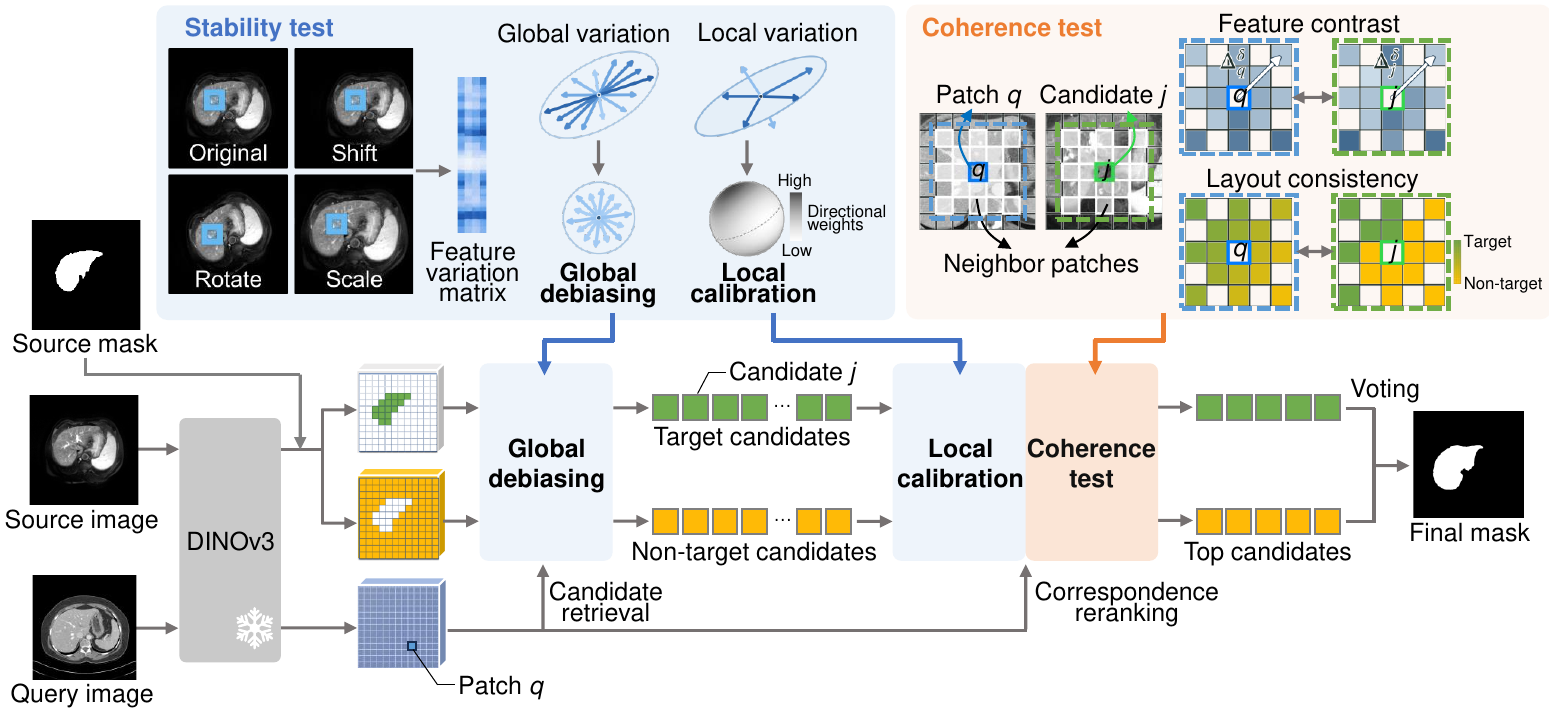}

    \caption{
    \textbf{Overview of CAST.}
    Given an annotated source image, its mask, and an
    unlabeled query image, CAST calibrates and refines
    source--query patch correspondences through two
    complementary tests.
    The stability test characterizes feature variations
    induced by geometric transformations, providing
    global debiasing before candidate retrieval and
    patch-specific metrics for subsequent reranking.
    The coherence test evaluates retrieved candidates
    through neighborhood feature contrast and
    target--non-target layout consistency.
    Within the correspondence refinement framework,
    globally debiased features guide class-conditional
    candidate retrieval, after which source-specific
    metrics and neighborhood evidence jointly rerank
    the candidates.
    The top-$N_{\mathrm{top}}$ candidates from each class
    are then aggregated through voting to produce
    the final segmentation.
    }

    \label{fig:method}
\end{figure*}

\section{Related Work}

\noindent\textbf{In-context segmentation.}
In-context segmentation (ICS) enables new segmentation
tasks to be specified through a few annotated exemplars
at inference time.
Painter~\cite{wang2023painter} and SegGPT~\cite{wang2023seggpt}
establish general-purpose visual prompting through
large-scale training, while subsequent methods incorporate
dense feature correspondences into learned segmentation
frameworks~\cite{meng2024segic}.
Another line leverages pretrained vision foundation
models (VFMs), such as SAM~\cite{kirillov2023sam},
for visual prompting and exemplar-guided segmentation
~\cite{zhang2024persam,liu2024matcher,zhang2024bridge}.
More recently, self-supervised vision transformers
~\cite{caron2021dino,oquab2024dinov2,simeoni2026dinov3}
have demonstrated strong semantic and spatial properties
in their patch representations, enabling training-free
segmentation through direct feature matching.
INSID3~\cite{cuttano2026insid3} combines positional
debiasing with region-level clustering of DINOv3 features,
while FoRIS~\cite{hu2026foris} progressively refines
foreground correspondences through purification
and consolidation.
FROST~\cite{park2026frost} models foreground and
background feature distributions through nonparametric
density estimation.
Other studies explore feature selection, representation
refinement, and background interference
~\cite{zakir2026fssdino,sun2026rsrm,gopal2026rebase}.
Despite these advances, reliable patch-level
correspondence remains challenging under geometric
and appearance variations, particularly when feature
similarity alone fails to capture surrounding structure.

\medskip\noindent\textbf{Few-shot biomedical image segmentation.}
The diversity of anatomical structures and imaging
conditions, together with limited expert annotations,
makes few-shot learning particularly relevant to
biomedical segmentation~\cite{dissanayake2026fewshot}.
UniverSeg~\cite{butoi2023universeg} and
Tyche~\cite{rakic2024tyche} learn to generalize across
segmentation tasks through training on large collections
of annotated medical images.
Subsequent work further explores improved support
selection and similarity-based context refinement,
explicit in-context learning, and broader medical
task coverage
~\cite{gao2025dualcheck,xie2026eicseg,zhao2026segmic,
wu2024oneprompt,wong2025multiverseg,guo2026kprism}.
Medical adaptations of pretrained segmentation models
provide another route to biomedical applications.
MedSAM~\cite{ma2024medsam} extends SAM to medical
images, while MAUP~\cite{zhu2025maup} explores
exemplar-based prompting and HERA~\cite{ma2026hera}
employs test-time adaptation for cross-domain
generalization.
These approaches have also been extended to
volumetric medical imaging
~\cite{gao2025iris,hu2025neuroverse3d,
zheng2024msfseg,gao2026mass}.
Despite these advances, existing approaches often
rely on extensive medical annotations for pretraining,
medical-specific adaptation, or specialized volumetric
representations.
In contrast, we leverage the transferable representations
of a frozen VFM for training-free biomedical
in-context segmentation, without additional
medical-specific pretraining or gradient-based adaptation.

\section{Method}
\label{sec:method}

Given $K$ annotated source exemplars
$\mathcal{S}=\{(I_s^k,M_s^k)\}_{k=1}^{K}$
and an unlabeled query image $I_q$, our goal is to
segment the corresponding structure without task-specific
training or gradient-based adaptation.
We employ a frozen DINOv3 ViT-L/16 encoder~\cite{simeoni2026dinov3}
to extract dense visual descriptors from source and query
images.
The final transformer block (B24) produces
$D=1024$-dimensional descriptors on a $64\times64$
patch grid from $1024\times1024$ inputs.
All descriptors are L2-normalized.

As illustrated in \cref{fig:method}, CAST establishes
reliable source--query correspondences through two
complementary tests.
The stability test characterizes feature variations
under geometric transformations, providing
global debiasing before candidate retrieval and
patch-specific metrics for subsequent reranking.
The coherence test evaluates retrieved candidates
through neighborhood feature contrast and layout consistency.
These tests are integrated into a correspondence
refinement framework: globally debiased features
guide class-conditional candidate retrieval, while
source-specific metrics and neighborhood evidence
jointly rerank the retrieved correspondences.
The top-ranked candidates from each class are then
aggregated through voting to produce the final
segmentation.
For volumetric inputs, source-derived calibration
statistics and matching references are shared across
the query volume, while individual slices are processed
independently and their predictions stacked in order.

\subsection{Stability Test: Feature Calibration}
\label{sec:stability}

Geometric transformations can alter pretrained visual
features even when the underlying image content is
preserved.
We exploit controlled transformations of source images
to characterize feature variability at global and
patch-specific levels.
These source-derived statistics are computed once
for a fixed support set and reused across queries,
amortizing the calibration cost.

\paragraph{Global debiasing.}
For each source image $I_s^k$, we apply a fixed set
of non-identity geometric transformations
$\mathcal{T}=\{T_t\}$, including translations,
rotations, and isotropic scaling.
Let $\mathbf{f}_{k,p}$ denote the original descriptor
at patch location $p$, and $\mathbf{f}_{k,t}$ the feature
map extracted from the $t$-th transformed image.
We align transformed features with the original patch
grid through backward warping with bilinear sampling.
The aligned residual is
$\mathbf{r}_{k,t,p}
=[\mathcal{A}_{T_t}(\mathbf{f}_{k,t})]_p
-\mathbf{f}_{k,p}$,
where $\mathcal{A}_{T_t}$ denotes geometric alignment.

We aggregate these residuals into an uncentered
global second-moment matrix:
\begin{equation}
\mathbf{G}
=
\frac{1}{|\mathcal{B}|}
\sum_{(k,t)\in\mathcal{B}}
\frac{1}{|\mathcal{P}_{k,t}|}
\sum_{p\in\mathcal{P}_{k,t}}
\mathbf{r}_{k,t,p}\mathbf{r}_{k,t,p}^{\top},
\label{eq:global_moment}
\end{equation}
where $\mathcal{B}$ indexes source--transformation
pairs and $\mathcal{P}_{k,t}$ contains their valid
patch correspondences.
Each pair contributes equally to the estimate.
We extract the leading $R=32$ eigenvectors of
$\mathbf{G}$, forming
$\mathbf{U}\in\mathbb{R}^{D\times R}$,
which capture the dominant transformation-sensitive
directions.

Global debiasing suppresses these shared variations
through a soft projection:
\begin{equation}
\mathbf{z}(\mathbf{f})
=
\operatorname{Norm}
\left(
\mathbf{f}
-
\alpha\mathbf{U}\mathbf{U}^{\top}\mathbf{f}
\right),
\label{eq:soft32}
\end{equation}
where $\alpha$ controls the suppression strength
and $\operatorname{Norm}(\cdot)$ denotes L2 normalization.
We apply the same projection to original source,
transformed source, and query descriptors.
The resulting globally debiased feature maps are
denoted by $\mathbf{Z}_s$ and $\mathbf{Z}_q$,
with patch descriptors $\mathbf{s}_{k,p}$ and
$\mathbf{z}_q$.

\paragraph{Local calibration.}
Global debiasing captures shared feature variations,
but individual source patches may respond differently
to geometric transformations.
We therefore characterize their variability separately
in the debiased feature space.
For each original source patch, the aligned residual is
$\mathbf{e}_{k,p,t}
=[\mathcal{A}_{T_t}(\mathbf{z}(\mathbf{f}_{k,t}))]_p
-\mathbf{s}_{k,p}$.
We aggregate these residuals into patch-specific
second-moment matrices and construct the corresponding
matching metrics:
\begin{equation}
\begin{aligned}
\mathbf{C}_{k,p}
&=
\begin{cases}
\displaystyle
\frac{1}{n_{k,p}}
\sum_{t\in\mathcal{V}_{k,p}}
\mathbf{e}_{k,p,t}\mathbf{e}_{k,p,t}^{\top},
& n_{k,p}>0,\\
\mathbf{0}, & n_{k,p}=0,
\end{cases}\\
\sigma^2
&=
\frac{1}{KD}
\sum_{k=1}^{K}
\frac{1}{|\mathcal{P}_k|}
\sum_{p\in\mathcal{P}_k}
\operatorname{tr}(\mathbf{C}_{k,p}),\\
\mathbf{A}_{k,p}
&=
\mathbf{I}+\frac{\mathbf{C}_{k,p}}{\sigma^2},
\end{aligned}
\label{eq:local_metric}
\end{equation}
where $\mathcal{V}_{k,p}$ contains the valid
transformed observations and
$n_{k,p}=|\mathcal{V}_{k,p}|$.
$\mathcal{P}_k$ denotes the source patches used
for calibration.
Each source patch is thus associated with a
patch-specific metric reflecting its observed
feature variability, which is used for subsequent
correspondence reranking.

\subsection{Coherence Test: Neighborhood Verification}
\label{sec:coherence}

Feature similarity alone may not distinguish a target
structure from visually similar neighboring anatomy.
The coherence test therefore evaluates retrieved
correspondences through two complementary forms of
neighborhood evidence: feature contrast and layout consistency.

\paragraph{Feature contrast.}
We compare feature variations around query patch $q$
and source candidate $j$ relative to their respective
center descriptors.
Let $\mathcal{O}$ denote the $24$ ordered offsets
formed by eight directions at radii
$r\in\{1,2,4\}$.
For each common valid offset
$\delta\in\mathcal{O}_{qj}$, we compute the
center-relative feature differences
$\boldsymbol{\Delta}_q^\delta
=\mathbf{z}_{q+\delta}-\mathbf{z}_q$
and
$\boldsymbol{\Delta}_j^\delta
=\mathbf{s}_{j+\delta}-\mathbf{s}_j$.
We measure their agreement through normalized
correlation:
\begin{equation}
\kappa_{qj}
=
\begin{cases}
\displaystyle
\frac{
\sum_{\delta\in\mathcal{O}_{qj}}
\langle
\boldsymbol{\Delta}_q^\delta,
\boldsymbol{\Delta}_j^\delta
\rangle
}{
\sqrt{E_q(q,j)E_j(q,j)}
},
&
\substack{
|\mathcal{O}_{qj}|>0,\\
E_q(q,j),E_j(q,j)>\tau
},
\\[8pt]
0, & \text{otherwise},
\end{cases}
\label{eq:correlation}
\end{equation}
where $E_q(q,j)$ and $E_j(q,j)$ denote the summed
squared magnitudes of the corresponding feature
differences over common valid offsets.
The context distance and its reliability are defined as
\begin{equation}
\begin{aligned}
d_{\mathrm{ctx}}(q,j)
&=
\max\left(
1-\operatorname{clip}(\kappa_{qj},-1,1),
\epsilon
\right),\\
\rho_{qj}
&=
\frac{|\mathcal{O}_{qj}|}{24}
\mathbb{I}\left(
E_q(q,j)>\tau,\,
E_j(q,j)>\tau
\right).
\end{aligned}
\label{eq:context_reliability}
\end{equation}
The reliability score $\rho_{qj}$ weights feature
contrast according to neighborhood coverage and
local feature variation during reranking.

\paragraph{Layout consistency.}
We further assess whether the target--non-target
arrangement around a source candidate agrees with
the initial query unary map.
The source mask provides target and non-target labels
for neighboring source patches, while the unary map
provides corresponding target scores in the query.
We compare these two forms of neighborhood evidence:
\begin{equation}
R_{qj}
=
\frac{1}{24}
\sum_{\delta\in\mathcal{O}_{qj}}
(2y_{j+\delta}-1)
(2u(q+\delta)-1),
\label{eq:role}
\end{equation}
where $y_{j+\delta}\in\{0,1\}$ denotes the source
target/non-target label, and $u(q+\delta)$ is the
initial target score of the corresponding query
neighbor.
The comparison rewards agreement at each relative
position: source target neighbors are expected to
have high query target scores, whereas source
non-target neighbors should have low scores.
Higher values of $R_{qj}$ therefore indicate stronger
agreement between the source and query layouts.

\subsection{Correspondence Refinement Framework}
\label{sec:refinement}

The two tests are integrated into a correspondence
refinement framework that first retrieves and then
evaluates source--query matches.
Globally debiased features identify class-specific
candidates, which are subsequently reranked using
the precomputed patch-specific metrics and
neighborhood evidence before voting.

\paragraph{Candidate retrieval.}
Each original source patch is stored in a matching
memory with its globally debiased descriptor,
target or non-target label, and source-specific
metric $\mathbf{A}_{k,p}$.
Geometrically transformed views are used only for
feature calibration, while original source patches
serve as matching references.
The source masks organize these references into
separate target and non-target candidate pools.

For each query patch $q$, we retrieve up to
$N_{\mathrm{cand}}$ nearest candidates from each class
using cosine similarity in the globally debiased
feature space, forming $\mathcal{C}_1(q)$ and
$\mathcal{C}_0(q)$.
For each class, we average the inverse cosine distances
of its $N_{\mathrm{init}}$ nearest candidates and
normalize the two class-specific scores to obtain
the initial target score $u(q)$, which provides
the query unary map for layout consistency.

\paragraph{Correspondence reranking.}
For each retrieved candidate $j$, we evaluate its
center-level feature discrepancy using the
source-specific metric from the stability test:
\begin{equation}
d_{\mathrm{loc}}(q,j)
=
\frac{1}{2}
\left[
(\mathbf{z}_q-\mathbf{s}_j)^{\top}
\mathbf{A}_j^{-1}
(\mathbf{z}_q-\mathbf{s}_j)
+
\sigma^2\log\det(\mathbf{A}_j)
\right],
\label{eq:local_distance}
\end{equation}
where $\mathbf{A}_j$ denotes the metric associated
with source candidate $j$.
The quadratic term adapts feature discrepancies
to candidate-specific variability, while the
log-determinant term accounts for its overall
uncertainty.

We combine this stability-based distance with
neighborhood evidence from the coherence test.
To align their numerical scales, we introduce a
query-dependent factor $\eta_q$ based on median
distances across the candidate set.
Let
$\mathcal{C}(q)=\mathcal{C}_0(q)\cup\mathcal{C}_1(q)$,
and let $\bar d_{\mathrm{ctx}}(q,j)$ use
$d_{\mathrm{ctx}}(q,j)$ when neighborhood evidence
is available and $d_{\mathrm{loc}}(q,j)$ otherwise.
The scale factor and final correspondence distance are
\begin{equation}
\begin{aligned}
\eta_q
&=
\frac{
\operatorname{med}_{j\in\mathcal{C}(q)}
d_{\mathrm{loc}}(q,j)
}{
\max\left(
\operatorname{med}_{j\in\mathcal{C}(q)}
\bar d_{\mathrm{ctx}}(q,j),
10^{-6}
\right)
},\\
d^*(q,j)
&=
\Big[
(1-\beta\rho_{qj})d_{\mathrm{loc}}(q,j)
+\beta\rho_{qj}\eta_qd_{\mathrm{ctx}}(q,j)
\Big]e^{-R_{qj}},
\end{aligned}
\label{eq:final_distance}
\end{equation}
where $\beta$ controls the contribution of feature
contrast.
The resulting distance combines patch-specific
feature reliability, neighborhood contrast, and
layout agreement.
We independently rerank target and non-target
candidates using $d^*(q,j)$ and retain the
$N_{\mathrm{top}}$ nearest candidates from each class.

\paragraph{Voting.}
The selected correspondences provide inverse-distance
evidence for the two classes.
We aggregate this evidence independently to obtain
class-specific scores and the final target probability:
\begin{equation}
\begin{aligned}
b_c(q)
&=
\frac{1}{|\mathcal{V}_c^{N_{\mathrm{top}}}(q)|}
\sum_{j\in\mathcal{V}_c^{N_{\mathrm{top}}}(q)}
\frac{1}{\max(d^*(q,j),\epsilon)},\\
p(q)
&=
\frac{b_1(q)}{b_1(q)+b_0(q)},
\qquad c\in\{0,1\},
\end{aligned}
\label{eq:prediction}
\end{equation}
where $\mathcal{V}_c^{N_{\mathrm{top}}}(q)$ contains
the selected reranked candidates of class $c$.
By normalizing the evidence within each class,
voting balances the contributions of target and
non-target correspondences.
The resulting target scores are bilinearly upsampled
to the original image resolution and thresholded
at $0.5$ to obtain the final segmentation.

\section{Experiments}

\newcommand{\benchhead}[2]{%
  \shortstack[c]{#1\\[-1pt]{\footnotesize\itshape #2}}}

\begin{table*}[t]
\centering
\caption{
Comparison of in-context segmentation methods on eight biomedical
benchmarks (Dice, \%, $\uparrow$).
All tasks use five annotated source exemplars ($K=5$).
Same/Cross denote within-/cross-marker transfer on DMBM;
C/M denote CT/MRI, with arrows indicating transfer directions.
2D Dice is computed per target image, while 3D Dice is computed
over the entire target volume.
Avg. is the equal-weight mean across datasets.
Gray marks UniverSeg results on datasets included in its pretraining
(CHAOS and AMOS22), and their contribution to Avg.
Best and second-best results are shown in
\textbf{bold} and \underline{underlined}, respectively.
}
\label{tab:main_results}

\small
\setlength{\tabcolsep}{3.85pt}
\renewcommand{\arraystretch}{0.98}

\begin{tabular}{@{}l*{11}{c}@{\hspace{4.5pt}}c@{}}
\toprule
\multirow{4}{*}{\textbf{Method}}
 & \multicolumn{5}{c}{\textbf{2D segmentation}}
 & \multicolumn{6}{c}{\textbf{3D segmentation}}
 & \multirow{4}{*}{\textbf{Avg.}} \\
\cmidrule(lr){2-6}
\cmidrule(lr){7-12}

 & \multirow[c]{3}{*}{\benchhead{Chest}{X-ray}}
 & \multirow[c]{3}{*}{\benchhead{REFUGE}{Fundus}}
 & \multirow[c]{3}{*}{\benchhead{GlaS}{Histology}}
 & \multicolumn{2}{c}{DMBM}
 & \multirow[c]{3}{*}{\benchhead{CTPelvic1K}{CT}}
 & \multirow[c]{3}{*}{\benchhead{LPBA40}{MRI}}
 & \multicolumn{2}{c}{CHAOS}
 & \multicolumn{2}{c}{AMOS22}
 & \\[-1pt]

 & & &
 & \multicolumn{2}{c}{\footnotesize\itshape Fluorescence}
 & &
 & \multicolumn{2}{c}{\footnotesize\itshape CT / MRI}
 & \multicolumn{2}{c}{\footnotesize\itshape CT / MRI}
 & \\[-1pt]

\cmidrule(lr){5-6}
\cmidrule(lr){9-10}
\cmidrule(lr){11-12}

 & & & & Same & Cross
 & & & C$\to$C & M$\to$C
 & M$\to$M & C$\to$M
 & \\
\midrule

\multicolumn{13}{@{}l}{%
\itshape Segmentation-trained or test-time adapted} \\[-1pt]

SegGPT~\cite{wang2023seggpt}
 & \textbf{95.40} & \underline{82.63} & 71.99
 & 72.82 & 67.93
 & 39.11 & 27.32
 & \underline{79.70} & 50.31
 & \underline{32.34} & 24.76
 & \underline{60.05} \\

UniverSeg~\cite{butoi2023universeg}
 & 91.16 & 72.36 & 58.17
 & 75.50 & 71.93
 & \underline{39.23} & 24.86
 & \textcolor{black!55}{76.23} & \textcolor{black!55}{61.33}
 & \textcolor{black!55}{30.49} & \textcolor{black!55}{20.16}
 & \textcolor{black!55}{56.70} \\

HERA~\cite{ma2026hera}
 & 92.79 & 40.71 & 83.40
 & 61.77 & 60.94
 & 32.71 & 12.72
 & 75.11 & \underline{71.47}
 & 25.91 & \underline{25.37}
 & 52.83 \\

\addlinespace[0.5pt]
\multicolumn{13}{@{}l}{%
\itshape Training-free: mask-supervised pretraining} \\[-1pt]

GF-SAM~\cite{zhang2024bridge}
 & 67.98 & 54.11 & 79.44
 & 26.27 & 26.30
 & 3.72 & 2.27
 & 32.66 & 26.57
 & 5.73 & 4.98
 & 33.60 \\

MAUP$^\dagger$~\cite{zhu2025maup}
 & 77.27 & 14.55 & 61.27
 & 50.61 & 49.40
 & 4.90 & 1.37
 & 49.89 & 39.08
 & 7.76 & 7.72
 & 32.70 \\

\addlinespace[0.5pt]
\multicolumn{13}{@{}l}{%
\itshape Training-free: self-supervised features} \\[-1pt]

INSID3~\cite{cuttano2026insid3}
 & 87.30 & 11.83 & 68.61
 & 44.00 & 43.54
 & 4.45 & 5.53
 & 34.89 & 34.51
 & 9.66 & 8.43
 & 33.16 \\

FoRIS~\cite{hu2026foris}
 & 93.35 & 33.48 & 80.91
 & 30.80 & 27.10
 & 6.13 & 5.20
 & 32.08 & 27.56
 & 6.43 & 5.41
 & 35.47 \\

FROST~\cite{park2026frost}
 & 91.83 & 79.28 & \underline{83.56}
 & \underline{78.87} & \underline{75.84}
 & 31.01 & \underline{28.06}
 & 70.07 & 58.16
 & 25.99 & 17.09
 & 59.59 \\

\textbf{CAST (Ours)}
 & \underline{95.09} & \textbf{84.58} & \textbf{83.76}
 & \textbf{83.48} & \textbf{81.69}
 & \textbf{60.01} & \textbf{44.88}
 & \textbf{88.44} & \textbf{81.51}
 & \textbf{48.61} & \textbf{37.81}
 & \textbf{72.39} \\

\bottomrule
\end{tabular}

\vspace{1pt}
\begin{minipage}{\textwidth}
\footnotesize
$^\dagger$ MAUP’s 2D results are adapted to the five-shot setting by majority voting over five single-support predictions.
\end{minipage}
\end{table*}

\begin{figure*}[t]
    \centering
    \includegraphics[width=\textwidth]{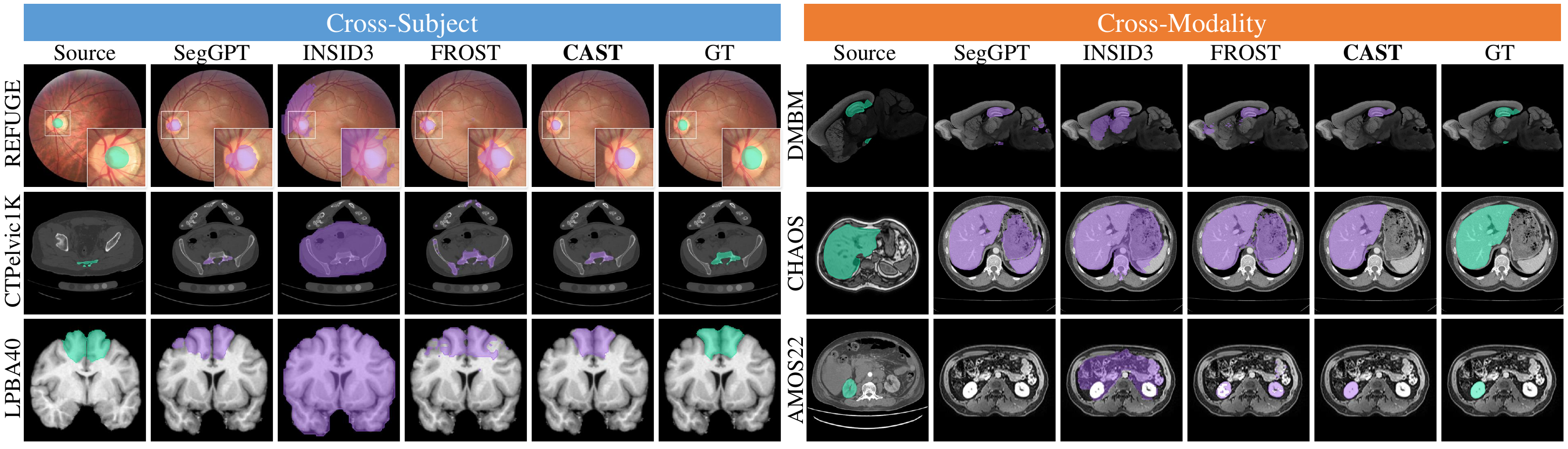}
    \caption{
    Qualitative comparison of CAST with SegGPT, INSID3, and FROST
    under the one-shot ($K=1$) setting.
    Left: cross-subject segmentation on REFUGE, CTPelvic1K,
    and LPBA40.
    Right: transfer across imaging appearance on DMBM (cross-marker),
    CHAOS and AMOS22 (cross-modality).
    Green overlays indicate reference annotations and ground truth,
    while purple overlays denote predicted masks.
    White boxes highlight the zoomed-in regions in REFUGE.
    }
    \label{fig:main_result}
\end{figure*}

\subsection{Experimental Setup}
\label{sec:experimental_setup}

We evaluate CAST on eight biomedical benchmarks spanning
2D and 3D segmentation under cross-subject and cross-modality transfer settings.
The 2D datasets include Chest X-ray~\cite{jaeger2014two}
(lungs), REFUGE~\cite{orlando2020refuge}
(optic disc and cup), GlaS~\cite{sirinukunwattana2017gland}
(glands), and DMBM~\cite{yuan2026prmix}
(11 brain regions).
DMBM contains fluorescence images acquired with
PSD95 and SAP102 markers, enabling both within-marker
and cross-marker transfer.
The 3D benchmarks comprise CTPelvic1K~\cite{liu2021deep}
(4 pelvic structures), LPBA40~\cite{shattuck2008construction}
(6 brain regions), CHAOS~\cite{kavur2021chaos}
(liver), and AMOS22~\cite{ji2022amos}
(6 abdominal organs).
For volumetric segmentation, we consider CT$\to$CT
and three MRI$\to$CT settings on CHAOS, and
MRI$\to$MRI and CT$\to$MRI on AMOS22.
Detailed dataset protocols and structure-wise results
are provided in the supplementary material.

Each episode consists of $K=5$ annotated source exemplars
and an unlabeled query image or volume.
We use fixed, randomly sampled episodes shared across
all methods, with distinct source and query images.
Chest X-ray, REFUGE, and GlaS use 600 episodes per setting.
For DMBM and the 3D benchmarks, we evaluate 100 episodes
per anatomical structure and transfer direction.
For 3D tasks, five annotated slices from one source subject
guide the segmentation of another subject's complete volume.
Each query slice is processed independently, and the
predictions are stacked to reconstruct the volume.

We report Dice (\%)~\cite{taha2015metrics}
per query image for 2D tasks and over the complete
query volume for 3D tasks.
Results are averaged equally across anatomical structures
and transfer directions, with the overall average assigning
equal weight to the eight datasets.
We compare CAST with segmentation-trained or adapted
methods (SegGPT~\cite{wang2023seggpt},
UniverSeg~\cite{butoi2023universeg},
and HERA~\cite{ma2026hera}),
SAM-based methods (GF-SAM~\cite{zhang2024bridge}
and MAUP~\cite{zhu2025maup}),
and frozen self-supervised feature-based methods
(INSID3~\cite{cuttano2026insid3},
FoRIS~\cite{hu2026foris},
and FROST~\cite{park2026frost}).

\medskip\noindent\textbf{Implementation details.}
CAST uses a frozen DINOv3 ViT-L/16 backbone with
$1024\times1024$ inputs and no gradient updates.
For stability calibration, we apply seven fixed
geometric transformations to each source image,
including translations, rotations, and isotropic scaling.
We set $\alpha=0.5$ and $\beta=0.25$.
For candidate retrieval, we retrieve up to
$N_{\mathrm{cand}}=100$ candidates per class and use
$N_{\mathrm{init}}=5$ nearest candidates per class
to construct the initial target unary map.
For correspondence reranking, we retain
$N_{\mathrm{top}}=5$ candidates per class, which are
then aggregated through voting.
Further implementation details are provided in the
Supplementary Material.

\subsection{Main Results}
\label{sec:main_results}

\Cref{tab:main_results} presents quantitative comparisons
across eight biomedical benchmarks, while
\cref{fig:main_result} provides representative
qualitative results.
CAST achieves the highest overall average Dice of
72.39\%, outperforming the strongest baseline,
SegGPT (60.05\%), by 12.34 percentage points.
The gains are particularly pronounced in volumetric
segmentation, where CAST ranks first across all
six settings.

\medskip\noindent\textbf{Cross-subject segmentation.}
Under same-modality transfer, CAST achieves strong
performance across diverse anatomical structures
and imaging conditions.
For 2D segmentation, it leads on REFUGE (84.58\%)
and GlaS (83.76\%), and achieves 83.48\% on DMBM
within-marker transfer, surpassing FROST by 4.61 points.
The gains become substantially more pronounced in
volumetric segmentation, where five annotated source
slices guide segmentation of an unseen 3D volume.
On CTPelvic1K and LPBA40, CAST achieves 60.01\%
and 44.88\% Dice, exceeding the strongest baselines
by 20.78 and 16.82 percentage points, respectively.
Similar improvements are observed on CHAOS CT$\to$CT
and AMOS22 MRI$\to$MRI, where CAST outperforms SegGPT by 8.74 and
16.27 points.
These results highlight the effectiveness of calibrated
patch correspondences for transferring sparse annotations
across subjects and complex anatomy, without requiring
mask-supervised pretraining or task-specific optimization.

\medskip\noindent\textbf{Cross-modality segmentation.}
Cross-modality transfer is particularly challenging
for appearance-based matching, as corresponding
anatomical structures may exhibit substantially
different contrasts and textures.
By incorporating feature stability and local structural
evidence, CAST maintains a clear advantage under
such appearance variations.
On CHAOS MRI$\to$CT and AMOS22 CT$\to$MRI,
it achieves 81.51\% and 37.81\% Dice,
outperforming HERA by 10.04 and 12.44 percentage
points, respectively.
Notably, HERA relies on test-time parameter adaptation,
whereas CAST achieves these improvements without
gradient-based updates. The advantage also extends to fluorescence images
acquired with different markers.
On DMBM, CAST achieves 81.69\% Dice in cross-marker
transfer, surpassing FROST (75.84\%) by 5.85 points.

\subsection{Further Analysis}
\label{sec:analysis}

\begin{figure}[t]
    \centering
    \includegraphics[width=0.95\columnwidth]{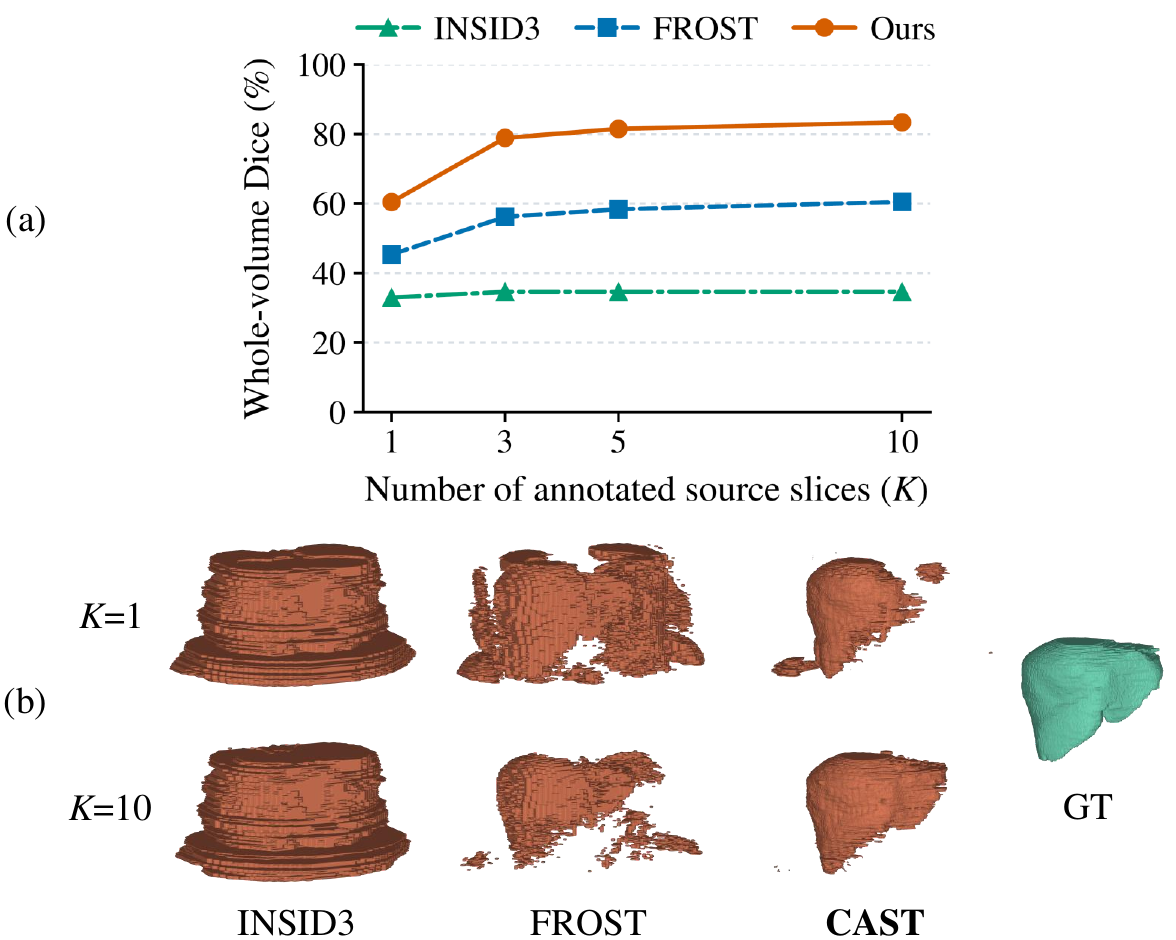}
    \caption{
    Effect of the source annotation budget on CHAOS
    MRI$\to$CT.
    Top: Whole-volume Dice of INSID3, FROST, and CAST
    with $K\in\{1,3,5,10\}$ annotated source slices.
    Bottom: Representative 3D liver segmentations
    at $K=1$ and $K=10$, with ground truth (GT)
    for reference.
    }
    \label{fig:k_shot}
\end{figure}

\noindent\textbf{Effect of the source annotation budget.}
To assess annotation efficiency, we compare CAST with
FROST and INSID3 under varying source annotation budgets
on CHAOS MRI$\to$CT.
As shown in \cref{fig:k_shot}, CAST consistently
outperforms FROST and INSID3 across all annotation
budgets.
With only one annotated slice, CAST achieves 60.47\%
Dice, compared with 45.35\% for FROST and 33.03\%
for INSID3.
Notably, CAST reaches 78.93\% with just three annotated
slices, already surpassing FROST's 60.52\% at $K=10$.
Its performance further increases to 81.51\% and
83.38\% at $K=5$ and $K=10$, respectively,
demonstrating its ability to achieve accurate segmentation with minimal annotation.

\medskip\noindent\textbf{Runtime comparison.}
We further examine the computational efficiency of CAST
on CHAOS MRI$\to$CT, using a dedicated NVIDIA A100 80GB GPU.
For INSID3 and FROST, we use source-reuse adaptations
of their official inference implementations.
As shown in \cref{tab:runtime}, CAST achieves the
lowest total runtime of 34.76\,s per volume, compared
with 74.48\,s for INSID3 and 65.54\,s for FROST.
Its faster target inference reduces the average runtime
to 0.184\,s per slice, including source setup.
This advantage is particularly valuable in practical
settings where the same annotated exemplars are reused
across multiple target volumes, as source calibration
only needs to be performed once.

\begin{table}[t]
\centering
\caption{
Runtime comparison on CHAOS MRI$\to$CT at $K=5$
(seconds).
Setup is performed once per annotated source set.
Total is the sum of setup and inference time per volume.
Avg. per slice includes amortized setup time.
}
\label{tab:runtime}

\small
\setlength{\tabcolsep}{3pt}
\renewcommand{\arraystretch}{1.1}

\begin{tabular*}{\columnwidth}{
    @{\extracolsep{\fill}}lcccc@{}
}
\toprule
\textbf{Method}
 & \textbf{Setup}
 & \textbf{Inference}
 & \textbf{Total}
 & \textbf{Avg. per slice} \\
\midrule

INSID3
 & \textbf{0.29}
 & 74.19
 & 74.48
 & 0.394 \\

FROST
 & 3.55
 & 61.99
 & 65.54
 & 0.347 \\

\textbf{CAST}
 & 7.01
 & \textbf{27.75}
 & \textbf{34.76}
 & \textbf{0.184} \\

\bottomrule
\end{tabular*}
\end{table}

\begin{table}[t]
\centering
\caption{
Ablation of CAST (whole-volume Dice, \%, $\uparrow$).
$\checkmark$ and -- indicate enabled and disabled tests.
All configurations use $K=5$.
CHAOS reports the mean across three MRI$\to$CT settings.
}
\label{tab:ablation}

\small
\setlength{\tabcolsep}{3pt}
\renewcommand{\arraystretch}{1.12}

\begin{tabularx}{\columnwidth}{
    p{0.12\columnwidth}
    *{4}{>{\centering\arraybackslash}X}
}
\toprule
 & \textbf{Stability}
 & \textbf{Coherence}
 & LPBA40
 & CHAOS \\
\midrule

\multicolumn{3}{l}{%
    \textcolor{black!55}{Direct label transfer (1-NN)}}
 & \textcolor{black!55}{37.72}
 & \textcolor{black!55}{77.53} \\

 & --
 & --
 & 41.09 & 79.10 \\

 & $\checkmark$
 & --
 & 42.54 & 79.44 \\

 & --
 & $\checkmark$
 & 42.58 & 79.67 \\

\textbf{CAST}
 & $\checkmark$
 & $\checkmark$
 & \textbf{44.88} & \textbf{81.51} \\

\bottomrule
\end{tabularx}
\end{table}

\medskip\noindent\textbf{Ablation study.}
We examine the contributions of the stability test
and the coherence test on LPBA40 and CHAOS, as shown
in \cref{tab:ablation}.
We also include direct 1-NN label transfer, which assigns
each query patch the label of its nearest source patch
in the uncalibrated feature space, to assess the
effectiveness of the overall matching framework.
The stability test alone achieves 42.54\% and
79.44\% Dice, while the coherence test alone reaches
42.58\% and 79.67\% on LPBA40 and CHAOS, respectively.
Combining both tests further improves performance
to 44.88\% and 81.51\%, with gains of 3.79 and
2.41 percentage points, respectively.
The additional improvements over either individual test
highlight the complementary roles of feature calibration
and neighborhood-based structural evidence in refining
anatomical correspondences.

\subsection{Challenging Application: Atlas Label Transfer}
\label{sec:atlas_transfer}

To further explore the potential of CAST in biomedical
research, we consider atlas label transfer, an important
application for bringing standardized anatomical
knowledge to specimen-specific images.
Brain atlases provide a common reference for localizing
molecular signals and comparing findings across
specimens and imaging modalities~\cite{wang2020allenccf}.
Transferring atlas annotations could reduce the need
for labor-intensive manual delineation, but remains
challenging due to substantial differences in imaging
appearance and the often indistinct boundaries between
neighboring brain regions~\cite{yuan2026prmix}.

\begin{figure}[t]
    \centering
    \includegraphics[width=0.9\columnwidth]{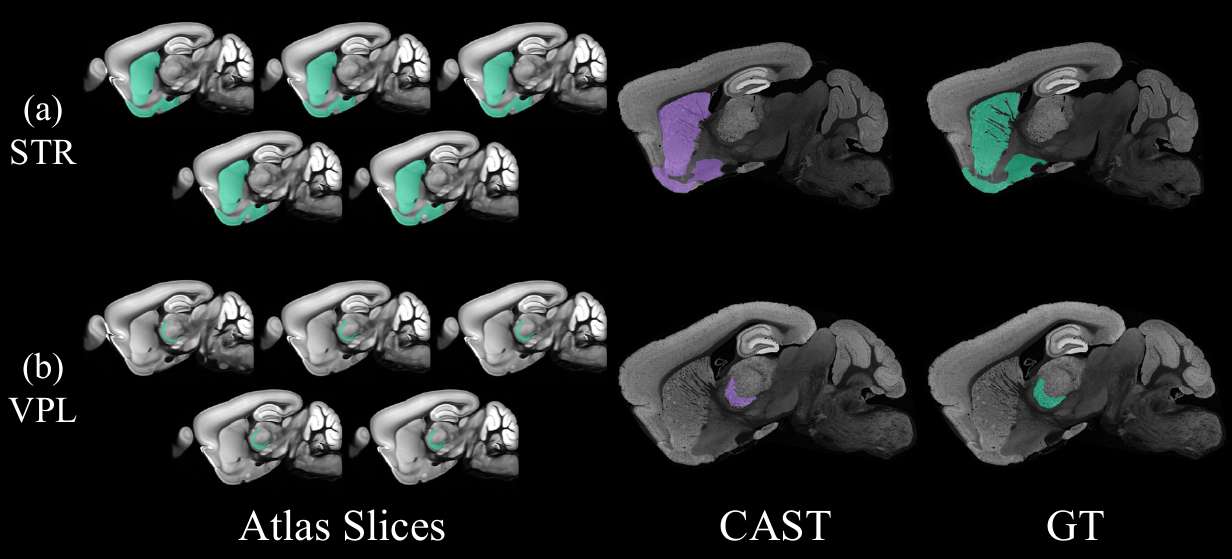}
    \caption{
    Qualitative atlas label transfer results in DMBM
    for the striatum (top) and ventral posterolateral
    (VPL) nucleus of the thalamus (bottom).
    }
    \label{fig:atlas_transfer}
\end{figure}

We evaluate CAST by transferring annotations from the
Allen Mouse Brain Common Coordinate Framework (CCFv3)
atlas~\cite{wang2020allenccf} to DMBM fluorescence images.
The evaluation covers 11 brain regions across PSD95
and SAP102, with 100 episodes per region--marker
combination. Each episode uses five annotated sagittal
atlas slices, with region-specific masks obtained
through ontology mapping.
As shown in \cref{tab:atlas_dmbm}, CAST achieves
66.92\% and 70.04\% Dice on PSD95 and SAP102,
respectively, yielding a mean of 68.48\% and
outperforming FROST by 6.83 percentage points.
The qualitative results in \cref{fig:atlas_transfer}
further illustrate CAST's ability to transfer atlas
labels without explicit registration, suggesting
promising applications in broader biomedical research.

\begin{table}[t]
\centering
\caption{
Atlas label transfer to DMBM using five annotated source slices
(Dice, \%, $\uparrow$).
Mean denotes the average of the PSD95 and SAP102 results.
}
\label{tab:atlas_dmbm}

\small
\setlength{\tabcolsep}{2.5pt}
\renewcommand{\arraystretch}{1.08}

\begin{tabular}{@{}lccccc@{}}
\toprule
& \multicolumn{5}{c}{\textbf{Method}} \\
\cmidrule(lr){2-6}
\textbf{Setting}
& SegGPT
& GF-SAM
& INSID3
& FROST
& \textbf{CAST (Ours)} \\
\midrule
PSD95
& 50.38 & 27.77 & 37.70
& \underline{59.73} & \textbf{66.92} \\

SAP102
& 50.05 & 27.43 & 44.89
& \underline{63.57} & \textbf{70.04} \\

\midrule
Mean
& 50.22 & 27.60 & 41.30
& \underline{61.65} & \textbf{68.48} \\
\bottomrule
\end{tabular}
\end{table}

\section{Conclusion}

We presented CAST, a training-free framework for
biomedical in-context segmentation that makes
frozen visual representations more reliable for
fine-grained anatomical matching.
Across eight biomedical benchmarks covering 2D and 3D
segmentation, cross-subject, and
cross-modality transfer, CAST improves average Dice
by 12.34 percentage points over the strongest baseline
and ranks first across all six volumetric settings.
Beyond conventional benchmarks, CAST achieves
68.48\% mean Dice in atlas label transfer, outperforming
the strongest baseline by 6.83 percentage points
when transferring standardized anatomical annotations
to specimen-specific microscopy images.
These results highlight the potential of CAST to
extend reliable segmentation beyond conventional
task-specific settings.
The versatility of CAST opens up broader opportunities
for annotation-efficient biomedical image analysis,
from 2D-to-3D segmentation to atlas-based applications.

\section*{Acknowledgements}

We thank Prof.~Sixian You for providing part of the computational
resources used in this work and for helpful discussions.

{
    \small
    \bibliographystyle{ieeenat_fullname}
    \bibliography{main}
}

\end{document}